\documentclass[11pt,a4paper]{article}
\usepackage{acl2015}
\usepackage{times}
\usepackage{latexsym}
\usepackage{comment}
\usepackage[colorinlistoftodos]{todonotes}

\usepackage{url}

\usepackage{graphics}      
\usepackage{subcaption}
\usepackage{verbatim}
\usepackage{color}
\usepackage{multirow}

\title{Benchmarking Popular Classification Models' Robustness to Random and Targeted Corruptions}

\author{Utkarsh Desai \\
  IBM Research \\
  \And
  Srikanth Tamilselvam \\
  IBM Research \\
  \And
  Jassimran Kaur \\
  IBM Research \\
  \And
  Senthil Mani \\
  IBM Research \\
  \And
  Shreya Khare \\
  IBM Research \\}
  
\date{}

\begin{document}
\maketitle
\begin{abstract}
Text classification models, especially neural networks based models, have reached very high accuracy on many popular benchmark datasets. Yet, such models when deployed in real world applications, tend to perform badly. The primary reason is that these models are not tested against sufficient real world ``natural" data. Based on the application users, the vocabulary and the style of the model's input may greatly vary. This emphasizes the need for a model agnostic test dataset, which consists of various corruptions that are natural to appear in the wild. Models trained and tested on such benchmark datasets, will be more robust against real world data. However, such data sets are not easily available. In this work, we address this problem, by extending the benchmark datasets along naturally occurring corruptions such as \textit{Spelling Errors}, \textit{Text Noise} and \textit{Synonyms} and making them publicly available. Through extensive experiments, we compare random and targeted corruption strategies using Local Interpretable Model-Agnostic Explanations(LIME). We report the vulnerabilities in two popular text classification models along these corruptions and also find that targeted corruptions can expose vulnerabilities of a model better than random choices in most cases.
\end{abstract}

\section{Introduction}
Building a machine learning model for any text based tasks such as classification, is difficult due to the vast vocabularies involved or generation task due to the difficulty in generating coherent and grammatically valid sentences. Unlike other domains such as computer vision, where a single value (pixel) may not have a high impact on the task, a change in a single word (or even a single character) in text can alter the class label or make the text grammatically incorrect ~\cite{Li2018TextBuggerGA}. In the past, heavy feature engineering was applied to produce an acceptable text based model ~\cite{Aggarwal2012ASO,survey1,url}. However, recently with the advent of neural networks, there has been a tremendous break-through in several text domain tasks such as sentiment analysis over user reviews and comments~\cite{MEDHAT20141093}, automatically detecting online toxic content ~\cite{toxicdata} such as irony, sarcasm among others and recommendations \cite{ziegler2005improving}. Yet, such models when deployed in real world scenarios perform poorly and behave in unexpected ways~\footnote{https://ind.pn/2CFnBlA}

The key problem for such a performance is the model's dependence on data. Unlike humans, these neural network models cannot generalize easily. Humans have the ability to analyze data comprising of various variations, corruptions and perturbations and hence can perform better than any existing systems. Humans are not muddled by variations like small typos, text length, sentence voice etc, as evident from this excerpt, about the cognitive process of reading text referred to as \textit{Typoglycemia}.\footnote{https://en.wikipedia.org/wiki/Typoglycemia} \\

\noindent\fbox{\begin{minipage}{19em}
\textit{Aoccdrnig to a rscheearch at Cmabrigde Uinervtisy, it deosn't mttaer in waht oredr the ltteers in a wrod are, the olny iprmoetnt tihng is taht the frist and lsat ltteer be at the rghit pclae. The rset can be a toatl mses and you can sitll raed it wouthit porbelm. Tihs is bcuseae the huamn mnid deos not raed ervey lteter by istlef, but the wrod as a wlohe}
\end{minipage}}\\

However, current neural network models need a huge number of examples comprising of all such variations as training data for a good, generalizable performance.

Given that many industries are starting to employ such models in their applications that interface with end customers, it has become critical to test the model's robustness to such variations with a comprehensive test dataset. However, collecting such relevant data is a huge challenge and if the variations are not adequately represented, the model might turn out to be sensitive to these variations and its performance might end up being over-estimated. 

One way to overcome this challenge is through data augmentation. Many recent works ~\cite{kobayashi-2018-contextual,wang-yang-2015-thats} have used random data augmentation techniques to increase the training set and have shown improvements on model's performance. However, our goal is not to improve the model's performance, but expose the model's vulnerabilities through a standard test set, consisting of natural variations. 

In this work, we argue that we can introduce small natural variations or \textit{corruptions} in the seed test set to increase its coverage. Thereby, providing a more realistic measure of the model's performance in a real world context when evaluated on the new test set. For example, consider a sentiment analysis model trained on a public forum data. If the model does not account for the fact that people often use abbreviations, incorrect grammar and even emojis in forum posts, it is bound to perform poorly when deployed. Hence, it is imperative that the model is tested for these natural variations. We employ techniques to introduce spelling errors, textual noise, and synonyms into a test set to thoroughly test the model's robustness. These corruptions can be introduced randomly at the sentence level, or using tools such as Local Interpretable Model-Agnostic Explanations (LIME) ~\cite{LIME} to guide the process. Further, to highlight the importance of such thorough testing, we perform an empirical study by benchmarking two high performing, popular text classification models against our generated test samples comprising of different forms of corruptions.

We conducted experiments on two classification models \textit{FastText} and \textit{Bidirectional LSTM} against four datasets \textit{SST2},  \textit{IMDB},  \textit{YELP}, and \textit{DBPEDIA}. Our experiments provide evidence that text classification models are indeed sensitive to corruption in the input and report lower accuracies on corrupted test sets compared to that on the uncorrupted, default test sets. Also, corrupting important words in a document produces a stronger drop in accuracy than randomly corrupting words.

Our contributions can be summarized as:

\begin{enumerate}
    \item Highlight the need for a corrupted benchmark test dataset consisting of natural variations for text based classification models. 
    \item Introduce spelling errors, text noise and synonyms and generate a corrupt test dataset from the given seed test dataset.
    \item Evaluate the value of LIME based and random strategies for introducing such corruption in the test dataset. 
    \item An empirical evaluation on the performance of popular text based classification models on natural variations introduced in the test set.
\end{enumerate}


The rest of the paper is organized as follows. We present an overview of various existing works in testing of deep learning models, data augmentation methods and other related work in Section 2. Our data corruption methods, strategies and the overall approach are presented in Section 3. We describe our experiments and discuss the results in Section 4. Finally, in Section 5, we summarize our findings and identify areas of future work.

\section{Related Work}

Algorithms have been developed to search for minimal distortions in input that generate an adversarial sample by both utilizing (white-box) and ignoring (black-box) model specifics. Works like \cite{ExplainingAH,adv_text,carlini1,Ebrahimi2017HotFlipWA}, have successfully generated adversarial perturbations for image data  \cite{Li2018TextBuggerGA,deeptext,Alzantot2018GeneratingNL,jia-liang-2017-adversarial} and text data respectively.

Additionally, research has also focused on gauging the testing adequacy of deep learning systems. \cite{testing1} propose a white-box framework for testing real-world deep learning systems using measures such as neuron coverage, trained on self-driving cars challenge data. Importance of a good quality test set is further emphasized by \cite{mutation_testing} using a white box mutation testing framework for image data. \cite{model_vulnerability} asses the vulnerability of computer vision models to distributional shifts on an input image. \cite{combi_testing} propose combinatorial testing criteria specialized for deep learning systems  

However, the fact that the performance of deep learning models is heavily coupled with its data is still largely ignored. Majority of studies assess model performance by computing accuracy on a standard test set, overlooking the representation of various variations in the test set. A deep learning model can be immune to many known types of adversarial attacks but may fail from unseen data variations. 

Furthermore, many recent works have incorporated data augmentation to improve  model performance.\cite{Mueller:2016:SRA:3016100.3016291} replaced random words in a sentence with their respective synonyms, to generate augmented data and train a Siamese recurrent network for sentence similarity task. \cite{wang-yang-2015-thats} used word embeddings of sentences to generate augmented data for the purpose of increasing data size and trained a multi-class classifier on tweet data.
\cite{kobayashi-2018-contextual} trained a bi-directional language model conditioned on class labels for generating augmented sentences wherein the words in the sentences were replaced by words having paradigmatic relation with the original word. \cite{Wei2019EDAED} suggest simple techniques such as synonym replacement, random insertion, random swap, and random deletion, for data augmentation to improve  performance of CNN and RNN models.

In contrast, the focus of our work is to test the models for natural variations, and we employ some of these techniques from prior art to generate such natural variations. We do not aim to generate adversarial examples but rather, generate test-cases with natural variations for assessing the model thoroughly. \cite{image_perturbation} benchmarks various image classification models for common image corruptions and surface variation and is a major inspiration for our work. To the best of our knowledge, we are the first to introduce corrupted benchmark test datasets for text using both random and LIME based strategies and asses the vulnerability of popular classification models for these corruptions. 





\begin{table*}[t]
\small
\centering
\begin{tabular}{llll}
\hline
\textbf{Methods} & \textbf{Level} & \textbf{Description} & \textbf{Example} \\
\hline
\hline
Missing & CHRL & A character is unintentionally dropped from & \textit{pr\textbf{o}blem} --\textgreater \\
Characters &  & a word. & \textit{prblem} \\
\hline
 Keyboard  & CHRL & A character can often be mis-typed incorrectly with  & \textit{probl\textbf{e}m}  --\textgreater\\
  Proximity&  & a neighbouring character on the keyboard (we consider  & \textit{probl\textbf{r}m}\\
   &  & only the QWERTY' keyboard layout) &    \\
\hline
 Adjacent  & CHRL & Adjacent characters in a word are interchanged & \textit{li\textbf{ke}ly} --\textgreater \\
 Character Swap &  &  accidentally &  \textit{li\textbf{ek}ly} \\
\hline
 Character & CHRL & A character is unintentionally pressed more than  & \textit{problem} --\textgreater \\
 Repetition &  & once &\textit{\textbf{pp}roblem} \\
\hline
 Homophones & WRDL & Incorrect usage of words having the same pronunciation & \textit{their} --\textgreater \textit{there} \\
&  & but different meanings, origins, or spelling. This type of & \textit{brake} --\textgreater \textit{break}\\
&  & typo is most likely to happen in speech to text conversion & \\
 \hline
\end{tabular}
\caption{Different character level and world level methods used for introducing spelling errors in text. }
\label{typostable}
\end{table*}

\begin{table*}[t]
\small
\centering
\begin{tabular}{llll}
\hline
\textbf{Methods} & \textbf{Level} & \textbf{Description} & \textbf{Example}  \\
\hline
\hline
Random  & CHRL & A random character is replaced by any & \textit{pr\textbf{o}blem} --\textgreater \\
Characters &  & random character. &  \textit{pr\textbf{x}blem} \\
\hline
 Special & CHRL & A random character is replaced by a special& \textit{prob\textbf{l}em} --\textgreater \\
 Symbols &  & symbol & \textit{prob\textbf{*}em}  \\
\hline
 Stop Words & WRDL & Common  stopwords such as \textit{the, is, at, which,  by, } & \textit{He fought fiercely}.\\
 &  &  \textit{and} are present unnecessarily. & --\textgreater \textit{He \textbf{is} fought fiercely} \\
\hline
 Whitespaces & CHRL & Unnecessary spaces between any characters & \textit{Wedding} --\textgreater \textit{We dding}\\
\hline
 Emojis & WRDL & Presence of emoji’s - extremely common in social  & \textit{I am very happy today}. \\
  &  &  media & --\textgreater \textit{I am very \textbf{:-D} today} \\
 \hline
 Homoglyphs & CHRL & characters, or glyphs with identical shapes,  & \textit{W\textbf{O}W!} --\textgreater \textit{W\textbf{0}W!}  \\
  &  & Again, very common in social media. &  \\
 \hline
\end{tabular}
\caption{Different character level and world level methods used for introducing noise in text}
\label{noise-table}
\end{table*}

\section{Approach}
Our goal is to introduce natural text corruptions to a test set and generate a new test set that is expected to have a closer resemblance to real-world data. To this end, we corrupt a dataset along multiple dimensions individually using multiple corruption methods and two corruption strategies. In this section, we first introduce our corruption methods, followed by an overview of the corruption strategies and finally our complete approach.

\subsection{Corruption Methods}
 We have employed three corruption methods, namely, introduction of spelling errors, noise addition and synonym replacement.  
 
\subsubsection{Spelling Errors}

With the amount of user-generated text data being exchanged on social media and other online channels, it is almost guaranteed that any datasets derived from these sources will exhibit human tendencies to make spelling errors. We account for two types of spelling errors:
\begin{enumerate}
    \item Typographical error (often abbreviated to typo) or spelling error that occur due to slips of the hand or finger while typing on electronic devices.
    \item Other spelling errors may be due to incorrect usage of words.
\end{enumerate}

Table ~\ref{typostable} list the different techniques we employ for introducing spelling errors in text that may occur at character (CHRL) or word level (WRDL).

\subsubsection{Text Noise}
Noise in text data refers to the presence of any unnecessary characters that reduce the overall readability of text. The type and frequency of text noise depends greatly on the data source and the data collection mechanism. The overall effect is still reduction in readability and comprehension of text. Interestingly, in most cases, humans are still able to understand text with high noise levels. Noise can be present both at character(CHRL) or word level (WRDL). Table ~\ref{noise-table} list the techniques used for introducing noise to corrupt the text data. 

\subsubsection{Word Synonyms}
One of the simplest ways to corrupt a given text document is to replace key words in the document with their synonyms. Depending on the selected words and their synonyms, this can drastically increase or decrease the readability of the document. Yet, it is almost guaranteed to preserve most of the document's overall meaning and thus it's class label. We generate a synonym set for a word using Wordnet~\cite{wordnet} and replace the word with a randomly selected synonym from this set. In our experiments, we only consider words classified as \textit{Nouns}, \textit{Adjectives}, \textit{Verbs} and \textit{Adverbs} for corruption via synonym replacement.

\begin{figure*}
\centering
\includegraphics[width=\textwidth]{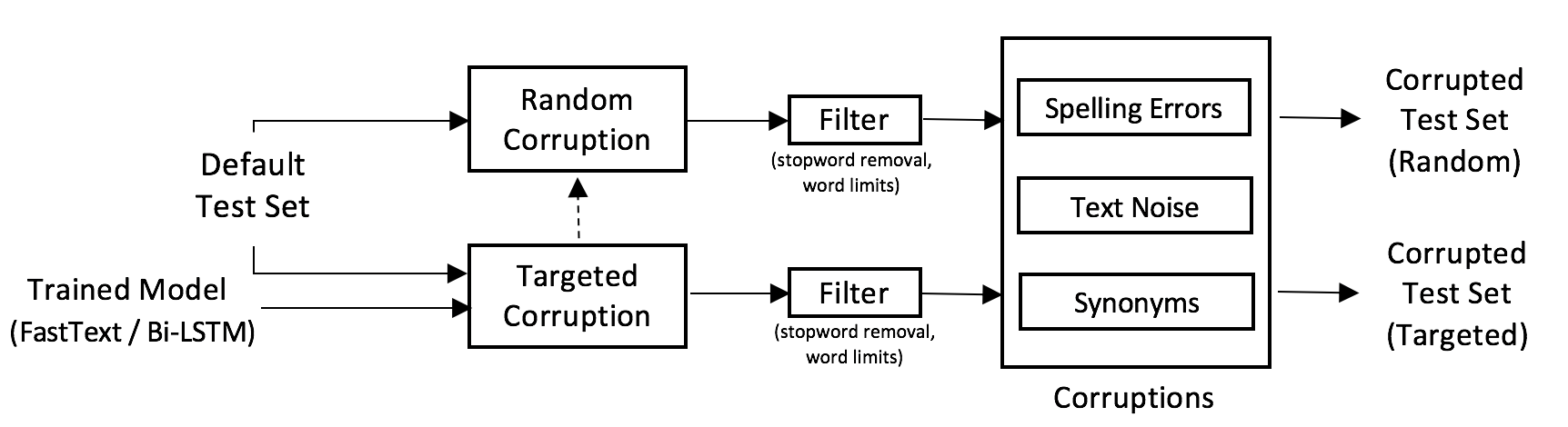}
\caption{Our transformation pipeline for introducing corruptions in a test set}
\label{fig:sys_arch}
\end{figure*}

\subsection{Corruption Strategies}

The most important decision in introducing corruptions is selecting the location of the change in the text. Instead of leveraging the internals of the model or any domain-specific knowledge about the data to identify the right words to be corrupted, we choose a model and data agnostic approach for general applicability. This leaves us with only a few options in selecting the locations within the document to introduce corruptions. We describe these strategies and provide our motivation for considering them below. 

\subsubsection{Random Corruption}
The simplest method to introduce corruptions to a document is to add them randomly. Any word in the document is eligible for a potential corruption using any of the techniques listed above. This is motivated by the fact that, by nature, all words in a document should be equally likely to get corrupted. All words are equally likely to be misspelled, noise can be present at any location in the document and any word may be expressed differently and yet, the document may still convey the same meaning. Thus, our first strategy is to randomly select words to be corrupted from the input document.

\subsubsection{Targeted Corruption}
Our goal is to ultimately test classification models. Therefore, an alternate hypothesis is that although all words in a document are equally likely to get corrupted, the result of corruptions would certainly be different for different words. Since, not all words are given equal importance by a model, this implies that words that are more important, should strongly affect the model's prediction upon corruption.

Local Interpretable Model-Agnostic Explanations (LIME)~\cite{LIME} is a popular black-box method to produce explanations of model predictions. For text inputs, the method returns a list of words in the document, ranked in decreasing order of importance, that were responsible for a particular prediction by the chosen model. In other words, the rank of a word can be interpreted as the importance of the word according to the model for a given prediction. Hence, our alternate scheme for selecting corruption targets is to prefer the words in the order generated by LIME. At first, this may seem to be equivalent to generating adversarial examples for text, except our goal is not to fool the classifier, but to produce enough variations in the test data.

\subsection{Complete Approach}
Our approach is best described as a pipeline of operations (Figure~\ref{fig:sys_arch}). First, the existing test dataset and the model being evaluated are provided as input. The only data cleaning we apply is to remove all punctuation from the input documents, except sentence delimiters.

Second, we apply both random and targeted strategies (using LIME) and select the words for corruption with just one additional condition - \textit{Stopwords} are not eligible for corruption. We argue that \textit{stopwords} convey very little information in the overall meaning of the document. Further, most models tend to give very little importance to \textit{stopwords} (explicitly or indirectly). Thus, in the Random Corruption strategy, we do not consider \textit{stopwords} for random selection. With Targeted Corruption, if LIME recommends a \textit{stopword} as having high importance, we ignore the word and pick the next important word. 


Finally, once the corruption targets are determined, we apply the desired corruption method and produce a corresponding corrupted test set.

\section{Experiments and Results}
In this section, we first discuss the details of the models and datasets used for our experiments. We then present our experiments and results for comparison of random and targeted corruptions strategies followed by the experiments and benchmarking results for the two models against the $4$ datasets.


\subsection{Datasets and Models}
For our experiments, we employ two widely used Text Classification models. The \textit{FastText} classifier is a skip gram model where the label is used as the middle word, followed by a softmax layer ~\cite{fasttext2,fasttext1}. The second model is a standard \textit{Bidirectional LSTM} (Bi-LSTM) with pretrained GloVe embeddings\cite{pennington2014glove}. 

We evaluate the robustness of these two models on $4$ datasets from varied domains and with different characteristics. Specifically, we use the \textit{SST2} \cite{socher-etal-2013-parsing}, \textit{IMDB} \cite{LMDB}, \textit{Yelp Reviews} \footnote{https://www.yelp.com/dataset/challenge} and \textit{DBPedia} \cite{dbpedia} datasets for the text classification task. \textit{SST2} is a two class movie review dataset with single sentence reviews. The \textit{IMDB} dataset is another movie review dataset with more documents and longer reviews. The Yelp polarity dataset was constructed from Yelp Dataset Challenge consisting of Yelp reviews with $1$(negative) or $5$(positive) ratings. The final dataset is the \textit{DBPedia} category classification dataset. The primary characteristics of these datasets are summarized in Table~\ref{table-dataset}.

\begin{table}[t]
\small
\centering
\begin{tabular}{lllll}
\hline
\textbf{Dataset} & \textbf{Test Set} & \textbf{Average}  & \textbf{Typos} & \textbf{Noise} \\ 
& \textbf{Size} & \textbf{Words/Doc} & \textbf{(\%)} & \textbf{(\%)} \\\hline
\hline
\textbf{SST2}    & 1700                   & 18  & 1.8 & 1.8                 \\ 
\textbf{IMDB}   & 25000                  & 233 & 2.9 & 2.3                \\ 
\textbf{YELP}    & 38000                  & 139 & 2.8 & 2.1                 \\ 
\textbf{DBPEDIA} & 70000                  & 51 & 10.7 & 8.3                    \\ \hline
\end{tabular}
\caption{Summary Statistics of the 4 datasets used in our experiments}
\label{table-dataset}
\end{table}

\begin{table}[t]
\small
\centering
\begin{tabular}{lll}
\hline
\textbf{Dataset}& \textbf{FastText} & \textbf{Bi-LSTM} \\ \hline
\hline
\textbf{SST2}    & 86.7              & 89.5             \\
\textbf{IMDB}    & 87.3              & 90.1             \\
\textbf{YELP}    & 93.1              & 96.6             \\
\textbf{DBPEDIA} & 96.7              & 99.0            \\
\hline
\end{tabular}
\caption{Baseline Accuracies of the FastText and Bi-LSTM models evaluated on the four datasets}
\label{table-models}
\end{table}

To be consistent across datasets and models, we apply minimal pre-processing to the data. All punctuation - except sentence delimiters - and HTML tags are removed from the documents and the text is then converted to lower case. We do not remove \textit{stopwords}. We then train the models with this pre-processed data. We use \textit{FastText} with default hyperparameters of vector size $10$, learning rate $0.1$ and wordNgrams $2$. We use a two layer \textit{Bi-directional LSTM} of hidden size $256$, learning rate $0.001$ and Adam optimizer, implemented using AllenNLP\cite{Gardner2017AllenNLP}. We train the two models on the default training sets for the four datasets and report the baseline performance of the models on the default test sets in Table~\ref{table-models}.

We perform two sets of experiments, one to validate LIME word rankings and the other to benchmark the selected models against corruptions. In all instances where we apply a random selection, the reported numbers are the mean values of three independent iterations. The eventual goal of our experiments is two fold:
\begin{enumerate}
\item To determine if one of the two corruption schemes clearly outperforms the other.
\item To determine the robustness of the two models against natural text corruptions.
\end{enumerate}

All datasets and experiment results are available here.\footnote{https://github.com/constraint-solvers/benchmark-corruptions}

\subsection{Selecting a LIME configuration}
We first aim to validate if LIME indeed selects words of higher importance. The explanation produced by LIME is conditioned on a lot of factors, including the model. So, different models may produce different explanations for the same document. To be able to correctly do so, we identify the optimal LIME configuration empirically. Thus, we first perform parameter selection for LIME.

The key factor that affects how accurately LIME identifies important words is the size of the neighborhood that is explored for each input document, controlled by a parameter \textit{num\_samples}. The size of the neighborhood depends on the number of words in the input document. Hence, we tune the value of the parameter \textit{num\_samples} individually for each dataset. The other important parameter is \textit{num\_features} that controls the number of words returned by LIME as part of its explanation. In all our experiments, we set \textit{num\_features} to be 15.

We examine $4$ values of the \textit{num\_samples} parameter: {750, 1500, 3000, 5000} using only the \textit{FastText} model. For each dataset, we construct a sample set consisting for randomly selected 10\% of the documents. In the case of \textit{SST2} dataset, we use 50\% of the documents since it has the lowest test data size among the datasets by far. For each value of \textit{num\_samples}, we identify the most important word according to LIME in each document of the four sample sets. We generate corresponding test sets where the most important word in the document is deleted. We compute the accuracies of these new test sets with the baseline \textit{FastText} model. For each dataset, we look for the value of \textit{num\_samples} that produces the highest drop in accuracy. The detailed results are presented in Table~\ref{table-lime}. 

\textbf{Results:}
We notice that the difference in accuracies is marginal across all datasets for different values of \textit{num\_samples}. However, the \textit{num\_samples} $750$ and $1500$  provide the most appropriate word rankings and higher values do not show a significant drop in the accuracy. We set \textit{num\_samples} to be $750$ in all our experiments.

\begin{table}[t]
\centering
\small
\begin{tabular}{llllll}
\hline
\textbf{Dataset} & \textbf{Baseline} & \multicolumn{4}{l}{\textbf{Sample Size (num\_samples)}}\\
 & \textbf{Accuracy} & \textbf{750} & \textbf{1500} & \textbf{3000} & \textbf{5000} \\
\hline
\hline
\textbf{SST2}  & 86.7 & 84.8         & 84.9          & 84.8          & \textbf{84.7}          \\
\textbf{IMDB}  & 87.3 & \textbf{85.0} & \textbf{85.0}          & 85.1          & 85.9          \\
\textbf{YELP} &  93.1  & 93.3         & 93.3          & \textbf{93.2}          & 93.4          \\
\textbf{DBPEDIA} & 96.7 & \textbf{95.6}& \textbf{95.6} & 95.7          & \textbf{95.6}         \\\hline
\end{tabular}
\caption{Test Set accuracies after deleting the top word in the LIME explanation for different values of num\_samples in comparison with the baseline accuracy}
\label{table-lime}
\end{table}




\subsection{Comparison of Strategies}
We now verify if targeted corruption using LIME is actually a valid alternative strategy to introduce corruptions randomly. We experimentally validate if the word importance ranking produced by LIME is indeed relevant. As before, we delete words from the documents and observe the change in test set accuracy. However, rather than deleting just one word from the document, we now use a parameter \textit{n} that controls the number of words to be deleted. Our argument behind this parameter is that for long documents, deleting a single word may not cause a change in the prediction because, the document would have multiple other words that carry some signal about the label. We vary the value of \textit{n} in \{$1$, $3$, $5$, $8$\}. 

\begin{figure*}
\centering
\includegraphics[width=\textwidth]{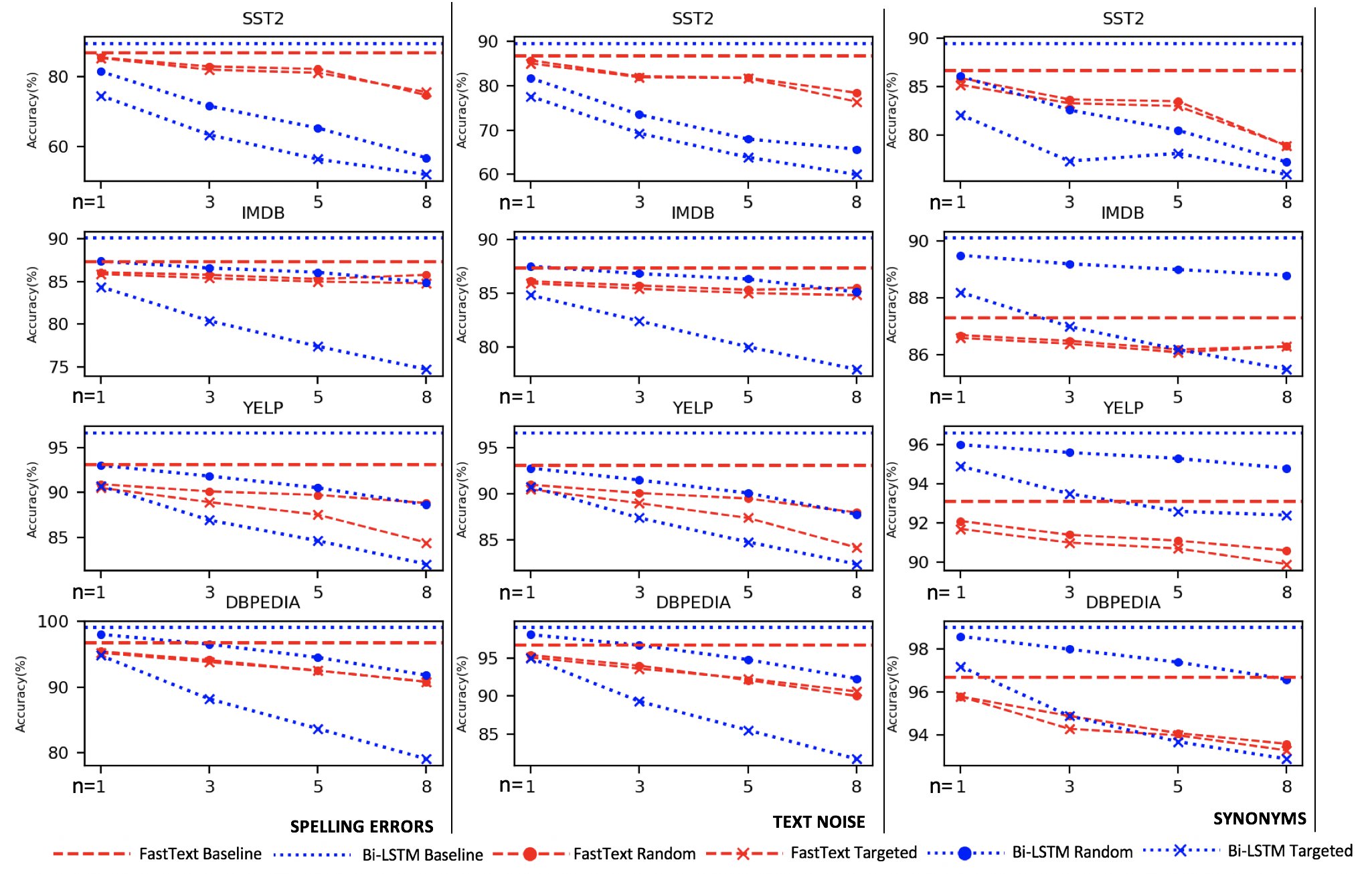}
\caption{Accuracies of the corrupted test sets on both models for different corruption strategies and corruption methods. The (red) dashed and (blue) dotted lines represent the FastText and Bi-LSTM models respectively. The horizontal lines are the baseline accuracies of the respective models. The Xs represent accuracy numbers using the targeted strategy and the Os represent accuracies using the random strategy.}
\label{fig:result_plot}
\end{figure*}

For a given test set, we first apply LIME to each document and obtain a ranked set of words, upto a maximum of $15$, which are important. We then remove any stopwords that may be present in this ranking and obtain a resulting set w\textsubscript{LIME}. If the number of words in this set is lesser than \textit{n}, we skip the document for that value of \textit{n}. If not, we also randomly select \textit{n} words from the document, ignoring stopwords, and obtain another set w\textsubscript{rand}. 

Now, from the original document, we thus generate two alternate documents by deleting the words from w\textsubscript{rand} and w\textsubscript{LIME}. We repeat this process for the test sets of each dataset and obtain $8$ alternate test sets, two for each dataset.  The accuracies of these test sets against their respective models are presented in Table~\ref{fasttext-accuracy} and ~\ref{lstm-accuracy}.

\begin{table}[t]
\small
\begin{tabular}{llllll}
\hline
\textbf{Dataset} & \textbf{Strategy} & \textbf{n=1} & \textbf{n=3} & \textbf{n=5} & \textbf{n=8} \\
\hline
\hline
\multirow{2}{*}{\textbf{SST2}} & Random & 85.3 & 81.8 & 81.4 & 76.4 \\
 & LIME & 84.8 & 81.4 & 80.5 & 71.4 \\
 \hline
\multirow{2}{*}{\textbf{IMDB}} & Random & 86.6 & 86.4 & 86.1 & 86.3 \\
 & LIME & 86.6 & 86.3 & 86.1 & 86.2 \\
 \hline
\multirow{2}{*}{\textbf{YELP}} & Random & 92.3 & 91.5 & 91.0 & 90.4 \\
 & LIME & 92.1 & 91.2 & 90.7 & 89.8 \\
 \hline
\multicolumn{1}{c}{\multirow{2}{*}{\textbf{DBPEDIA}}} & Random & 95.9 & 94.5 & 92.5 & 89.2 \\
\multicolumn{1}{c}{} & LIME & 95.8 & 94.4 & 92.2 & 90.1 \\
\hline
\end{tabular}
\caption{Accuracies on the FastText model for all datasets after deleting \textbf{n} words as per the Strategy}
\label{fasttext-accuracy}
\end{table}

\begin{table}[]
\small
\begin{tabular}{llllll}
\hline
\textbf{Dataset} & \textbf{Strategy} & \textbf{n=1} & \textbf{n=3} & \textbf{n=5} & \textbf{n=8} \\
\hline
\hline
\multirow{2}{*}{\textbf{SST2}} & Random & 84.7 & 81.9 & 80.0 & 76.2 \\
 & LIME &  77.0 & 70.1 & 70.1 & 70.0 \\
 \hline
\multirow{2}{*}{\textbf{IMDB}} & Random & 89.5 & 89.5 & 89.0 & 88.0 \\
 & LIME & 87.1 & 84.0 & 82.0 & 80.8 \\
 \hline
\multirow{2}{*}{\textbf{YELP}} & Random & 96.0 & 95.3 & 94.7 & 94.1 \\
 & LIME & 93.8 & 90.3 & 88.7 & 87.5 \\
 \hline
\multicolumn{1}{c}{\multirow{2}{*}{\textbf{DBPEDIA}}} & Random & 98.5 & 97.6 & 96.3 & 95.0 \\
\multicolumn{1}{c}{} & LIME & 96.3 & 91.6 & 88.4 & 86.5\\
\hline
\end{tabular}
\caption{Accuracies on the Bi-LSTMs model for all datasets after deleting \textbf{n} words as per the Strategy}
\label{lstm-accuracy}

\end{table}

\textbf{Results}: We notice that, in general across both models, test sets generated using the words from LIME explanations as corruption targets indeed produce lower accuracies than by randomly selecting words. Also, the drop in accuracies of these test sets increases with the increase in the number of words corrupted. As expected, deleting more words from a document reduces the accuracy on the test set, which can be seen for increasing values of \textit{n} compared to the baseline accuracies. 

Further, deleting words that have been selected by LIME as important produce a greater reduction in accuracy than deleting a random word. This difference is significant on the \textit{SST2} dataset  (upto $6.2\%$ for n=$8$), but minor for the other datasets. Referring to Table~\ref{table-dataset}, we can see that deleting upto $8$ words from an average sentence length of $19$ in \textit{SST2} would not preserve enough information in the document for an accurate classification. Since, we leave out stopwords, this would mean the classifier would have to make a prediction from even fewer informative words. The other datasets do not suffer from such a problem because the average document length is relatively high and there should be multiple words to make up for the information loss due to the deleted word. Although the difference between the accuracies on the test sets generated by deleting words in w\textsubscript{LIME} and w\textsubscript{rand} is not large, especially on the \textit{IMDB} dataset,  compared to the baseline accuracy of the default test sets, these drops are still significant. 

\subsection{Benchmarking against Corruptions}

The purpose of our second set of experiments and the primary contribution of this work is to benchmark the \textit{FastText} and \textit{Bidirectional LSTM} models against corruption on the $4$ datasets introduced earlier. We use a setup similar to the one described above for word deletion. That is, for a given document we obtain two sets of words, w\textsubscript{LIME} and w\textsubscript{rand} controlled by the parameter \textit{n}. Except, instead of deleting the selected words, we introduce corruptions using the techniques described in section $3.1$. 

For each dataset and each parameter value \textit{n}, we generate the sets w\textsubscript{LIME} and w\textsubscript{rand} for each document in their respective test sets using the two corruption strategies - random and targeted as described above. Then, we apply each of the three corruption techniques based on the words in w\textsubscript{LIME} and w\textsubscript{rand} on the respective documents. In summary, for each dataset, we apply LIME separately based on one of the two models. The output of LIME is filtered according to the parameter \textit{n} and a corresponding random set of words is also generated. Finally, the three corruptions are applied individually to words in w\textsubscript{LIME} and w\textsubscript{rand}. This gives a total of $192$ corrupted test sets, $96$ per model. A comprehensive visualization of the results of our experiments is presented in Figure~\ref{fig:result_plot}. 

\textbf{Results}: First, as expected, the accuracies of both models fall as the number of corrupted words in a document increases. This decrease in accuracy when compared to the original baselines is much greater on the \textit{SST2} dataset for both models (upto $11.1\%$ and $37.9\%$ for n=$8$). In general, the \textit{Bi-LSTM} model seems to have much larger drops in accuracy than the \textit{FastText} model (the blue dotted lines have higher slopes in all the plots than the red dashed lines). This is a very strong signal that the \textit{Bi-LSTM} model used in our experiments is highly vulnerable to corruptions. 

Additionally, both the models react strongly to the introduction of spelling errors and text noise. We believe this is because these two methods could potentially change the semantics of a document, while replacing words with synonyms is expected to preserve the original meaning. One minor anomaly we observe, specifically for synonym-based corruption, is that the accuracy increases as we corrupt more words (for \textit{SST2} dataset, \textit{Bi-LSTM} accuracy increases by $4.8\%$ and for \textit{IMDB} dataset, \textit{FastText} accuracy increases by $0.2\%$). We believe this increase is likely due to the replaced synonym was able to better convey the sentiment of the document. These experiments provide enough evidence that the two models we considered are vulnerable to the $3$ corruptions and could perform poorly if exposed to such corrupted data in production. The actual accuracy numbers corresponding to Figure \ref{fig:result_plot} and additional details on the impact of different types of spelling errors and text noise are available in the Appendix.

\section{Conclusion and Future Work}
Through extensive experiments, we successfully highlighted the need for benchmarking models for robustness against common corruptions. We also showed that targeted corruption strategies using LIME work better than random corruption. Even though our corruption strategies were simple, they were still able to reveal the vulnerabilities of the popular classification models. As future work, we aim to introduce more corruption strategies such as variations in paraphrases, voice forms and text length using sophisticated techniques such as GANs. We plan to benchmark more models against additional datasets. All benchmarked datasets and experiment results are made public.

\bibliography{emnlp-ijcnlp-2019}
\bibliographystyle{acl}

\end{document}